\documentclass[11pt,a4paper]{article}
\usepackage[hyperref]{acl2021}
\usepackage{times}
\usepackage{latexsym}

\usepackage{microtype}
\usepackage{graphicx}
\usepackage{multirow}
\usepackage{multicol}
\usepackage{amsmath}
\usepackage{cases}
\usepackage{array}
\usepackage{algorithm}
\usepackage{algpseudocode}

\aclfinalcopy 

\title{Dynamic Sliding Window for Meeting Summarization}

\author{Zhengyuan Liu, \ Nancy F. Chen \\
  Institute for Infocomm Research, A*STAR, Singapore \\
  \texttt{\{liu\_zhengyuan,nfychen\}@i2r.a-star.edu.sg}}

\date{}

\begin{document}
\maketitle
\begin{abstract}
Recently abstractive spoken language summarization raises emerging research interest, and neural sequence-to-sequence approaches have brought significant performance improvement. However, summarizing long meeting transcripts remains challenging. Due to the large length of source contents and targeted summaries, neural models are prone to be distracted on the context, and produce summaries with degraded quality. Moreover, pre-trained language models with input length limitations cannot be readily applied to long sequences. In this work, we first analyze the linguistic characteristics of meeting transcripts on a representative corpus, and find that the sentences comprising the summary correlate with the meeting agenda. Based on this observation, we propose a dynamic sliding window strategy for meeting summarization. Experimental results show that performance benefit from the proposed method, and outputs obtain higher factual consistency than the base model.
\end{abstract}

\section{Introduction}
\label{introduction}
Text summarization is studied in two paradigms: extractive and abstractive. Different from extractive models which directly select text spans from source content, abstractive approaches can generate summaries more concisely with flexible information aggregation and paraphrasing, but raise the higher requirement for context understanding and language generation. 
In the past few years, various neural sequence-to-sequence models are proposed and applied to abstractive summarization, from the vanilla attention-based summarizer \cite{rush-etal-2015-neuralSumm}, pointer-generator networks \cite{see-2017-pgnet}, to the recent large-scale pre-trained language models such as BERT and BART \cite{devlin-2019-BERT,lewis-etal-2020-bart}. These models have brought significant improvement on the generation quality especially in fluency and readability \cite{liu-lapata-2019-PreSumm}. 

While most prior work focuses on documents such as news articles \cite{hermann-2015-cnnDaily}, summarizing conversations starts to raise more research interest \cite{goo2018-gatedDiaSumm,zhu2020-SummHNet}. One typical multi-party conversation scenario is meeting \cite{mccowan2005ami}. Since meetings often aim to cover several sub-topic discussions, the average number of their dialogue turns is much larger than that of short conversations such as social chat \cite{gliwa-etal-2019-samsum} and inquiring-answering \cite{liu-2019-topic}. 
As the source content becomes longer, the summarizer may fall distracted on the massive contextual information, and the quality of neural generation tends to degrade when producing long text \cite{fan2018StoryGeneration}. Thus it is challenging for neural models to achieve the overall high quality for the long transcript summarization. Moreover, if adopting Transformer-based pre-trained language models like BART \cite{lewis-etal-2020-bart}, the transcript length often exceeds their maximum positional embedding limitations, and conducting text truncation will cause loss of some context.

Meetings usually consist of a set of agenda items \cite{mccowan2005ami}, and each item focuses on discussing a specific topic. Thus the whole meeting conversation is inherently organized at the topic level. This is also reflected in the summaries written by humans, which condense the key information on each discussion part. 
Therefore, we postulate the divide-and-conquer strategy can be useful to tackle the aforementioned challenges of meeting summarization. More specifically, our target is to split the long transcript into multiple segments, and obtain the final output by aggregating the summary snippets of each segment. To construct the ground-truth data for training, we first investigate the linguistic characteristics of meeting transcripts and reference summaries on the representative corpus AMI \cite{mccowan2005ami}.
Then, we propose to enhance the sequence-to-sequence summarizer with a dynamic sliding window strategy, to tackle the length limitation of Transformer-based language models. Unlike the conventional fixed window sliding \cite{devlin-2019-BERT}, the dynamic method makes a model learn to decide the start position of the next segment during the generation process.
Experimental results show that the summary generation can benefit from adopting the dynamic window sliding, and achieves state-of-the-art performance. Further example analysis demonstrates that the enhanced framework performs better than the base model considering factual consistency.

In the contemporary studies, efficient attention schemes of the Transformer \cite{beltagy2020longFormer} and multi-block context aggregation \cite{grail2021globalizing} are proposed to support long text processing, and \citet{schuller2020windowing} investigate the fixed and automatic sliding window for processing long documents, and they focused on summarizing Wikipedia and news articles. Our proposed method is inspired by the organizational features of meeting transcripts, and is applicable for the language backbones with input length limitations.

\begin{figure}
    \begin{center}
    \includegraphics[width=0.49\textwidth]{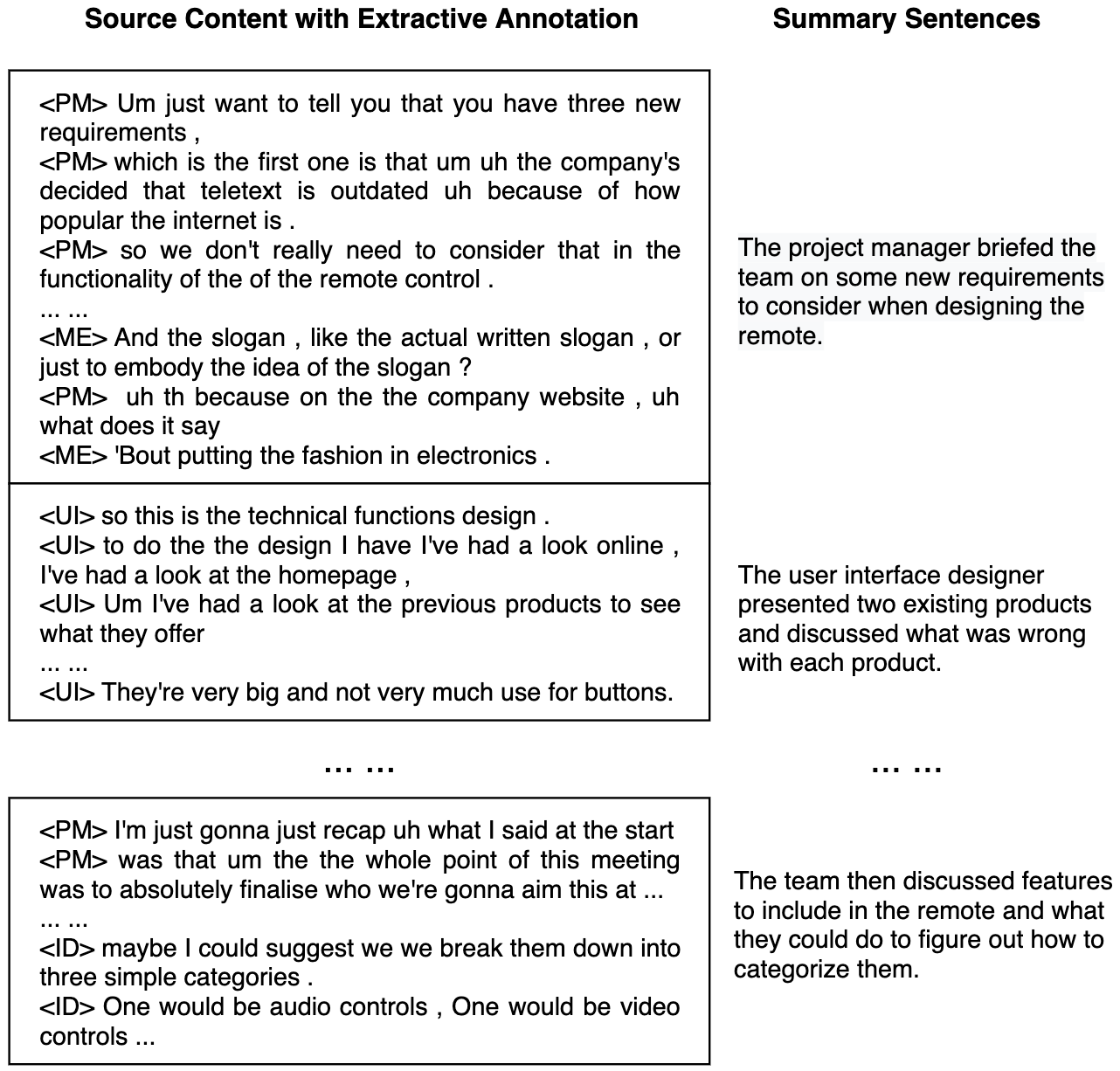}
    \end{center}
    \caption{One example of the meeting transcript and reference summary in the AMI corpus. Here we only show several extractive utterances in three sub-topic chunks and their corresponding abstractive sentences.}
    \label{fig:meet-example}
\vspace{-0.3cm}
\end{figure}

\begin{table}
\linespread{1.1}
    \centering
    \small
    \begin{tabular}{lcccc}
    \hline
         & CNN & DailyMail & NYT & AMI \\
    \hline
         \multicolumn{4}{l}{Average Source Content Length:} \\
         Sentence Level & 33.98 & 29.33 & 35.55 & 288.7 \\
         Word Level & 760.5 & 653.3 & 800.1 & 4757 \\
    \hline
    \multicolumn{4}{l}{Average Reference Summary Length:} \\
        Sentence Level & 3.59 & 3.86 & 2.44 & 17.55 \\
         Word Level & 45.8 & 54.65 & 45.54 & 323.3 \\
    \hline
    \end{tabular}
    \caption{Data statistics of the news summarization benchmarks and the AMI meeting corpus.}
    \label{tab:dataset}
\vspace{-0.2cm}
\end{table}

\section{Meeting Transcript Analysis}
\label{sec:meeting_analysis}
In this section, we conduct data analysis on the AMI meeting corpus \cite{mccowan2005ami}, in which the participants work in a team and conduct meetings to discuss product design, development, and planning. There are four main speaker roles: a project manager (PM), a marketing expert (ME), an industrial designer (ID), and a user interface designer (UI). Following previous work \cite{shang-etal-2018-unsupervised}, we split the whole dataset into train (97 transcripts), validation (20 transcripts), and test (20 transcripts) sets. The data statistics are shown in Table \ref{tab:dataset}, where we count one utterance in the conversation as one sentence, and conduct word-level tokenization. Compared with the news summarization benchmarks, the average length of meeting transcripts as well as the reference summaries are much larger (In our settings, we use the long version abstracts in the AMI corpus, as previous work \cite{feng2020-DDAGCN}). Moreover, human-written summaries of news articles often concentrate on the first few parts of source content \cite{liu2020conditional}, thus the truncation with a fixed length does not affect the final performance significantly \cite{jung2019earlier}. However, summarizing meetings requires grasping useful contextual information across the whole conversation, in this case, simple text truncation will lead to certain information loss.

\begin{figure}
    \begin{center}
    \includegraphics[width=0.43\textwidth]{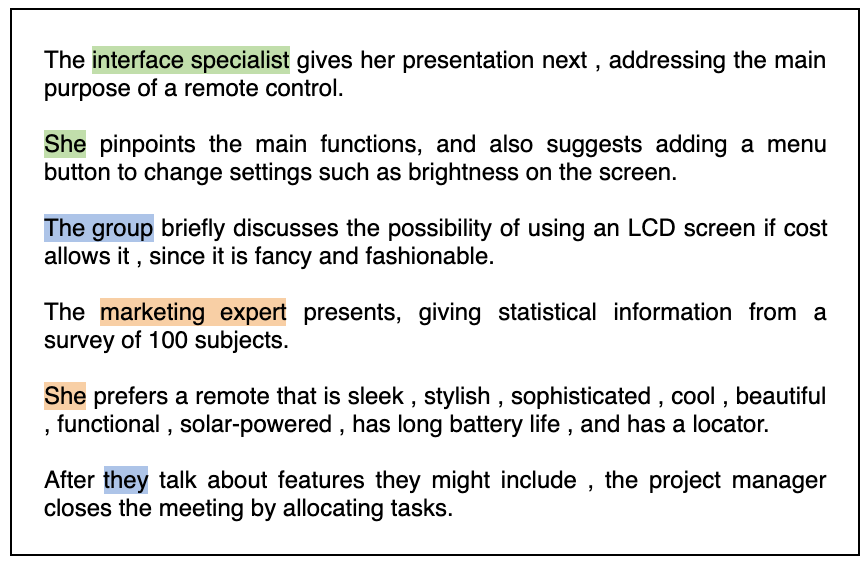}
    \end{center}
    \caption{One example of meeting summary in the AMI corpus. The colored spans are named entities or pronouns in a coreference chain.}
    \label{fig:coref-example}
\vspace{-0.3cm}
\end{figure}

\begin{figure*}[ht!]
    \begin{center}
    \includegraphics[width=0.82\textwidth]{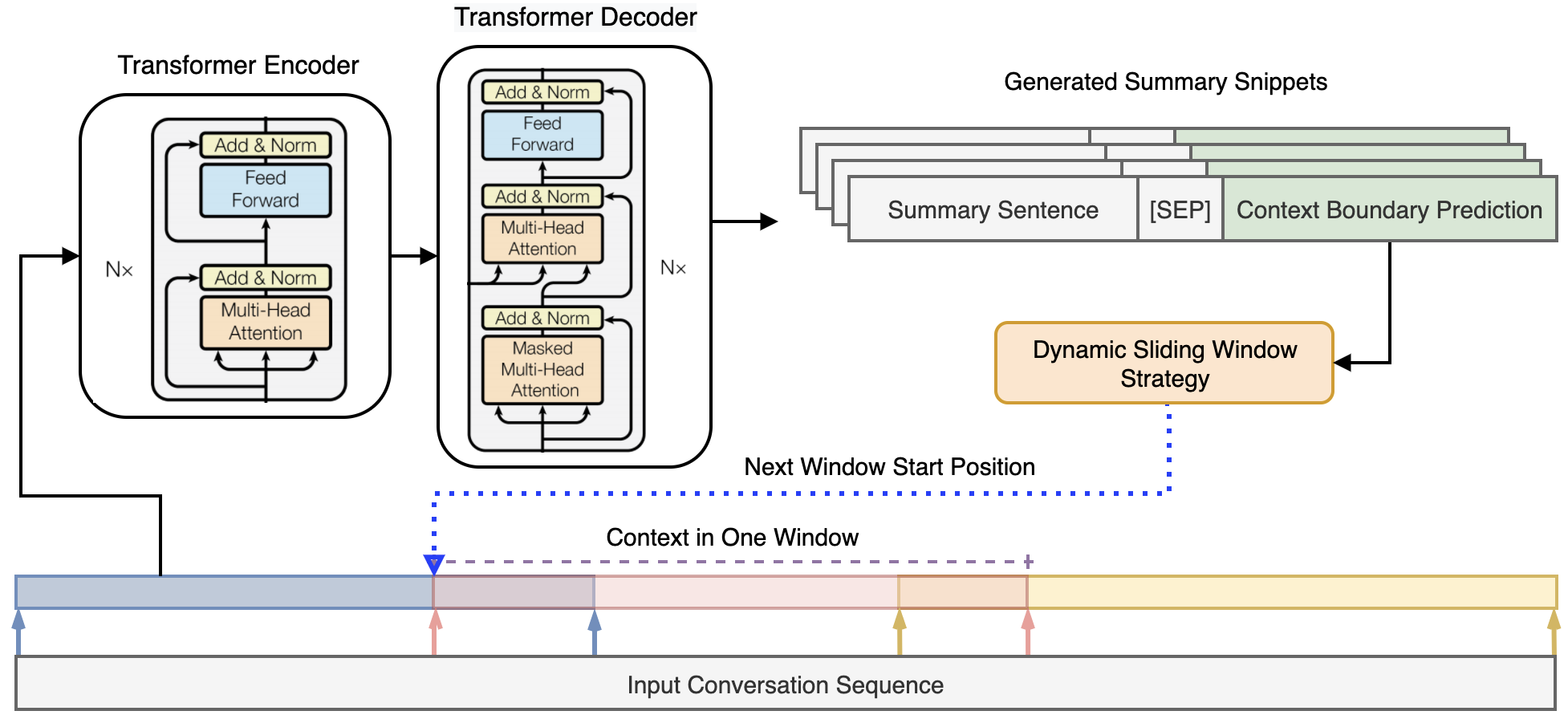}
    \end{center}
    \caption{Overview of the meeting summarization framework with the dynamic sliding window strategy.}
    \label{fig:framework}
\vspace{-0.2cm}
\end{figure*}

\subsection{Organization Analysis}
\label{ssec:organization_analysis}
Since meetings are usually taken with a set of sub-topics, we analyze the organizational co-relation of these sub-topics and sentences in meeting summaries.
In the AMI corpus, for each meeting transcript, aside from a human-written abstractive summary, there is an additional extractive annotation. As shown in Figure \ref{fig:meet-example}, human annotators were asked to choose sentences from the source conversation, as the supporting segments for each abstractive summary sentence. Based on this extractive annotation, we obtained the start/end supporting utterance indices in the conversation of each summary sentence. Then we sorted them by their occurrence order in the source content, and observed that 80\% sentences in reference summaries match the same agenda order as in the source content. This indicates that when humans write a summary, they often read the source content and record the key points sequentially.

\subsection{Segmentation Analysis}
\label{ssec:segment_analysis}
Summarization is a process to write a shorter version of a text span, and context integrity is essential. When adopting a divide-and-conquer strategy, only a part of the conversation will be extracted. To assess information integrity, we calculate the word level coverage with the extractive annotations (Figure \ref{fig:meet-example}) in the AMI. With the start/end supporting utterance indices described in Section \ref{ssec:organization_analysis}, we split the original transcripts into a number of segments. Then the word-level recall is calculated between each summary sentence and its corresponding segment. Summaries are often written via paraphrasing and introducing novel words, and the overall word-level coverage in our observation is 74\%.

\subsection{Summary Conversion}
\label{ssec:summary_conversion}
As the final output under a divide-and-conquer strategy is produced by aggregating summary snippets, to build the training ground-truth, we split each reference summary into multiple parts based on the analysis in Section \ref{ssec:organization_analysis} and Section \ref{ssec:segment_analysis}. For each summary sentence, we constructed its context segment by using the corresponding extractive annotation. Then we ordered the summary sentences by their occurrence indices in the source conversation, and merged adjacent summary sentences to summary snippets if their context segments had a certain overlap. For summary sentences without extractive annotation, we reorganized them via calculating word-level overlap with existing content segments.
Furthermore, we observed that some summary snippets are started with a pronoun that refers to a precedent personal named entity, as shown in Figure \ref{fig:coref-example}. To maximize the semantic integrity when producing one snippet, we used coreference resolution on the original summary, and if one summary snippet is started with a personal pronoun, it will be replaced by its referring.

\section{Neural Meeting Summarizer}
In this section, we elaborate our framework for meeting summarization with the dynamic sliding window strategy.

\subsection{Dynamic Sliding Window}
The sliding window is to split a long input into a number of shorter spans, process the spans in order (e.g., from left to right), and adopt aggregation for final output. It is a straightforward but useful method that is commonly applied to long sequence encoding \cite{devlin-2019-BERT}, and it is generally controlled by two parameters: \textit{window size} denotes the context span size, and \textit{stride size} is the amount of movement at each sliding step.
In previous work, these two parameters are often fixed, here we propose a dynamic sliding window strategy, where the stride size at each sliding step is predicted by the model. More specifically, given the window size $k$, at each sliding step, one index $j$ is selected (equals stride size is $j$) in the range of $[0, k-1]$ as the start position of next window.

\subsection{Base Neural Architecture}
For the neural summarizer, we use a Transformer-based sequence-to-sequence architecture \cite{vaswani-2017-Transformer}, and select the large-scale pre-trained language backbone BART \citep{lewis-etal-2020-bart} for initialization. The encoder consists of 6 stacked Transformer layers, and each layer has two sub-components: a multi-head self-attention layer and a position-wise feed-forward layer. Between the two sub-components, residual connection and layer normalization are used. The $u$-th encoding layer is formulated as:
\begin{equation}
\label{eq-attn-head}
    \widetilde{h}^u=\mathrm{LN}(h^{u-1}+\mathrm{MHAttention}(h^{u-1}))
\end{equation}
\vspace{-0.5cm}
\begin{equation}
\label{eq-attn-residual}
    h^u=\mathrm{LN}(\widetilde{h}^u+ \mathrm{FFN}(\widetilde{h}^u))
\end{equation}
where $h^0$ is the first layer input. $\mathrm{MHAttention}$, $\mathrm{FNN}$, $\mathrm{LN}$ are multi-head attention, feed-forward, and layer normalization, respectively.

The decoder consists of 6 stacked Transformer layers as well. In addition to the two encoding sub-components, the decoder performs another multi-head attention over the previous decoding hidden states and all the encoded representations. Then, the decoder generates tokens in an auto-regressive manner from left to right.

\subsection{Adopting Dynamic Sliding Window for Meeting Summarization}
\label{ssec:adopting_sliding}
The proposed dynamic sliding window is a general design that can be applied to various neural architectures.
In our summarization setting, the source content is a sequence with $n$ tokens $C=\{w_1, w_2,..., w_n\}$, and the summary is composed of $m$ snippets $S = \{s_1, s_2,..., s_m\}$. In a traditional sequence-to-sequence process, the whole source content $C$ is fed to a model as input, and the output summary is generated at one decoding stage. With the dynamic sliding window strategy, the encoding-decoding process will be conducted in a number of steps, as shown in Figure \ref{fig:framework}. Given the window size $k$, at $i$-th sliding step, the start token index is $i_{left}$, and the end token index $i_{right}$ is $i_{left} + k$. Then the input sequence of encoder at $i$-th step is $c_i=\{w_{i_{left}},..., w_{i_{right}}\}$.
After encoding, the contextualized hidden representation $h_i=\{h_{i_{left}},..., h_{i_{right}}\}$ is fed to the decoder to generate a summary snippet, and all snippets will be merged to form the final output.

To obtain the dynamic sliding prediction, one way is to directly predict the value of stride size as in \cite{gong-etal-2020-recurrentChunk}, where reinforcement learning is used to decide the stride value, and \citet{schuller2020windowing} proposed to generate a special indicator for the moving operation. Here, based on the analysis in Section \ref{sec:meeting_analysis}, we introduce a learning-efficient method named \textit{Retrospective}. More specifically, as shown in Figure \ref{fig:framework}, at $i$-th step, the decoder will not only generate the summary snippet $s_i$, but also the last supporting utterance of its context segment that is described in Section \ref{ssec:organization_analysis} (concatenated with a ``[SEP]'' token). Since the predicted supporting utterance provides the context boundary of the current generation step on the right, it is used to determine the amount of sliding movement. After all sliding steps, the generated snippets are concatenated as the output summary.

\section{Experiment Results and Analysis}

\subsection{Configuration}
The proposed framework was implemented using PyTorch \citep{paszke2019pytorch} and Hugging Face \citep{wolf-etal-2020-transformers}. The learning rate was set at 2e-5, and AdamW \cite{loshchilov2017-adamW} optimizer was applied. We trained each model for 20 epochs, and selected the best checkpoints on the validation set based on ROUGE-2 score \cite{lin-2004-rouge}. All input sequences were processed with the sub-word tokenization scheme as in \cite{lewis-etal-2020-bart}, and repeated output sentences were removed.
Based on the summary conversion described in Section \ref{ssec:summary_conversion}, we obtained 598/131/143 snippet-level samples as the train, validation, and test set, respectively. In the training stage, to simulate the input noise during inference time, we randomly added $k$ adjacent utterances ($5<k<15$) in each context chunk.
In the testing stage, for each example, the generation process of our \textbf{BART-SW-Dynamic} model was started with the first chunk of a default window size, which is initialized as 1024.\footnote{The token-level maximum input length of BART \cite{lewis-etal-2020-bart} is 1024 by default.} Then we used the sliding prediction described in Section \ref{ssec:adopting_sliding} to form the next chunk, until the whole transcript was processed. We also assessed the \textbf{BART-SW-Fixed} model by fixing the window size at 1024, the \textbf{BART-Truncate} model by truncating the transcript of 1024 tokens, and the \textbf{BART-SW-GoldSeg} model by using the gold context segmentation.

\begin{table}[t!]
\linespread{1.0}
\centering
\begin{tabular}{p{3.2cm}ccc}
\hline
                & \textbf{R-1} & \textbf{R-2} & \textbf{R-L}  \\
\hline
\multicolumn{2}{l}{Extractive baselines*:} \\
TextRank   &  35.19   &  6.13 &  15.70  \\
SummaRunner   &  30.98  &  5.54 &  13.91  \\
\hline
\hline
\multicolumn{2}{l}{Abstractive baselines*:} \\
Seq2Seq+Attention  &  36.42   &  11.97 &  21.78 \\
Pointer-Generator   &  42.60   &  14.01 &  22.62  \\
Sentence-Gated   &  49.29   & 19.31  &  24.82  \\
TopicSeg   &  51.53   & 12.23  &  25.47  \\
HMNet   & 53.02 & 18.57 & 24.85 \\
DDA-GCN  & 53.15  & 22.32 & 25.67 \\
\hline
\hline
\multicolumn{3}{l}{Our models:} \\
BART-SW-GoldSeg & 53.42  & 22.57 & 28.52 \\
BART-Truncate & 48.31  & 17.52 & 20.13 \\
BART-SW-Fixed & 50.02  & 19.86 & 23.98 \\
BART-SW-Dynamic & 52.83  & 21.77 & 26.01 \\
\hline
\end{tabular}
\caption{\label{output-ROUGE-table} ROUGE F1 scores on the AMI test set from baseline models and our framework. * Results of baselines are reported as in \cite{feng2020-DDAGCN}.}
\vspace{-0.2cm}
\end{table}

\subsection{Results on the AMI Corpus}
Following previous work \cite{chen-yang-2020-MultiView,feng2020-DDAGCN}, we used the ROUGE score \citep{lin-2004-rouge} for generation assessment, and reported ROUGE-1 (R-1), ROUGE-2 (R-2) and ROUGE-L (R-L) scores. We selected a set of strong baseline models for extensive comparison including Pointer-Generator \cite{see-2017-pgnet}, Sentence-Gated \cite{goo2018-gatedDiaSumm}, TopicSeg \cite{li-etal-2019-keepMeeting}, HMNet \cite{chen-yang-2020-MultiView}, and DDA-GCN \cite{feng2020-DDAGCN}. 
As shown in Table \ref{output-ROUGE-table}, for meeting summarization, abstractive approaches generally perform much better than extractive ones. In our settings, the model with the dynamic sliding window strategy (BART-SW-Dynamic) outperforms the model with the fixed window (BART-SW-Fixed), and the model with text truncation (BART-Truncate). This shows the effectiveness of the sliding window approach. Moreover, our proposed framework achieves the comparable performance of the contemporary state-of-the-art models.

\subsection{Sample Analysis}
\begin{figure}[t!]
    \begin{center}
    \includegraphics[width=0.48\textwidth]{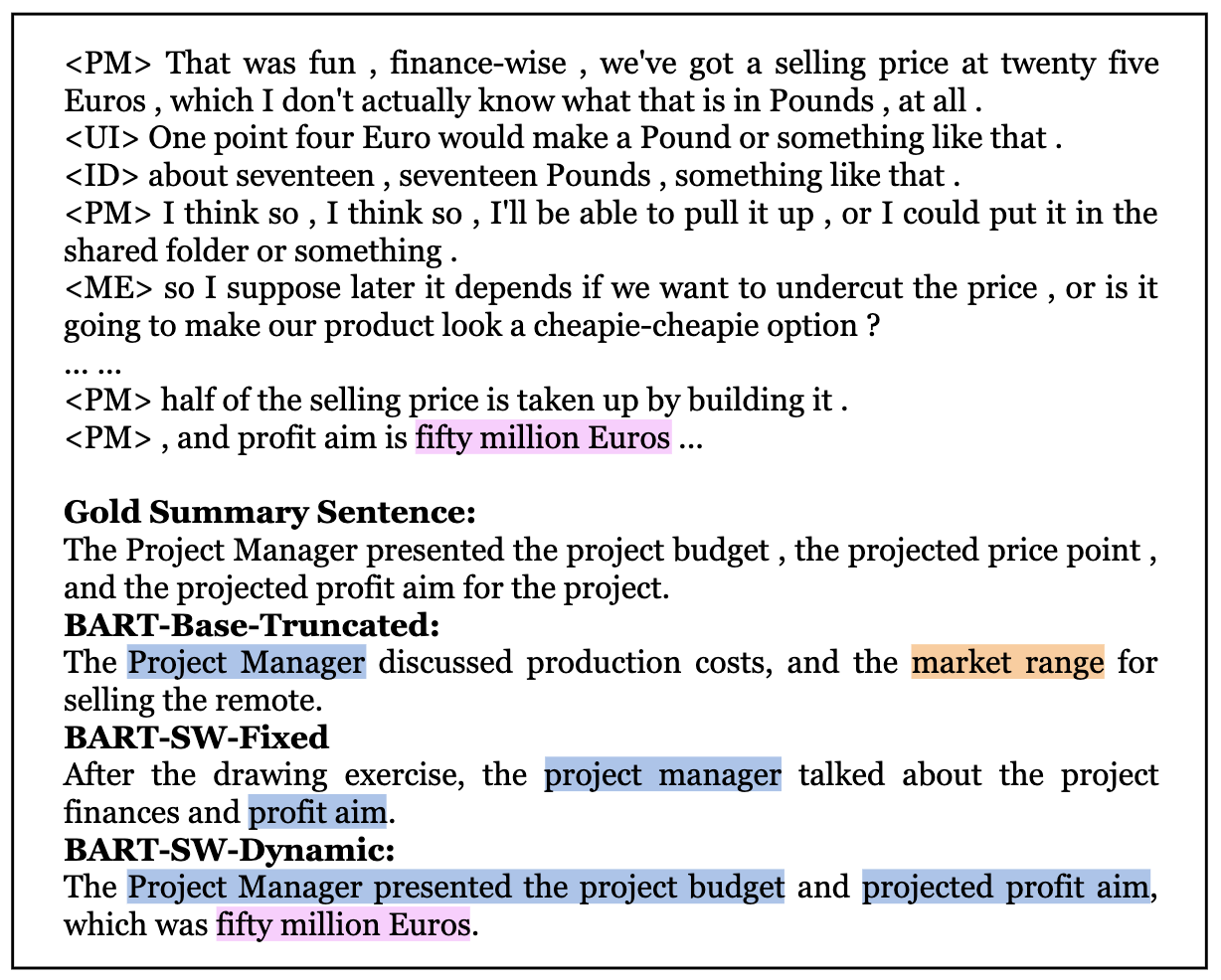}
    \end{center}
    \caption{Generated summaries. Spans in blue and pink are consistent with the gold summary and source content. Spans in yellow are factually inconsistent.}
    \label{fig:compare-example}
\vspace{-0.3cm}
\end{figure}

We first conducted a text quality analysis on the generated summaries across models. As shown in Figure \ref{fig:compare-example}, based on the strong generation capability of the language backbone, all BART-based models can produce fluent and grammatically correct summaries. While the model with text truncation achieves relatively acceptable ROUGE scores (as reported in Table \ref{output-ROUGE-table}), it produces sentences that are factually inconsistent with the source content, as shown in Figure \ref{fig:compare-example}. We speculate that this is caused by over-fitting the training samples. In contrast, since our proposed framework can produce the final summary based on relevant context segments without truncation, it performs better than the base model considering the factual consistency. 

We then conducted a stride prediction assessment. For the \textit{Retrospective} method described in Section \ref{ssec:adopting_sliding}, the predicted context boundaries are expected to be located closely to the ground-truth. With the best checkpoint, we observed that the average utterance-level distance between gold boundary span and model prediction is 2.7 (48 characters), and this shows that the model is able to predict the correct start position at each sliding step.

\section{Conclusion}
In this paper, to tackle the challenges from lengthy meeting transcript inputs, we proposed a dynamic sliding window strategy for abstractive summarization. Experimental results demonstrate that the neural sequence-to-sequence models can benefit from the proposed method, and suggest that the long transcript summarizing can be conducted in a divide-and-conquer manner. One of the future works can be extending the proposed method to other corpora with larger sample sizes.



\bibliographystyle{acl_natbib}
\bibliography{acl2021}


\end{document}